\documentclass[journal]{IEEEtran} 
\usepackage{graphicx}
\usepackage{epsfig}
\usepackage{bm}
\usepackage{booktabs}
\usepackage{xcolor}
\usepackage{listings}
\lstdefinestyle{terminal}{ backgroundcolor=
\color{gray!15}
, basicstyle=\ttfamily\small, frame=single, framerule=0pt, xleftmargin=0pt, xrightmargin=0pt,
breaklines=true, showstringspaces=false, keepspaces=true, numbersep=5pt }
\usepackage{times}
\usepackage{amsmath}
\usepackage{amssymb}
\usepackage{booktabs}
\usepackage{float}
\usepackage{algpseudocode}
\usepackage{algorithm}
\usepackage{balance}
\usepackage{multirow}

\definecolor{MyBlue}{HTML}{1E90FF}
\usepackage[
    urlcolor=MyBlue, 
    citecolor=MyBlue, 
    linkcolor=MyBlue, 
    colorlinks=true,
]{hyperref}

\title{\LARGE \bf
Interpreting Control Latents for System Identification \\via Conditional Flow Matching }

\author{Dingqi Zhang$^{*,1}$, Ruiqi Zhang$^{*,1}$, and Mark W. Mueller$^{1}$
\thanks{$^{1}$The authors are with the High Performance Robotics Lab, Dept. of Mechanical
Engineering, UC Berkeley. Contact at \{dingqi, richzhang, mwm\}@berkeley.edu. $^{*}$Equal
contribution.}
}

\begin{document}
    \maketitle

    \begin{abstract}
        Latent-conditioned adaptive policies can control robots across changing
dynamics, but their learned latents remain internal representations of the
policy rather than physical models that can be inspected, rolled out, or used
by other control modules. This limits closed-loop analysis, diagnosis, and
further improvement of a fixed policy. A direct mapping from latent to
physical parameters is also under-specified, because multiple systems can
induce similar closed-loop behavior. We therefore decode each operational
latent into a distribution of quadrotor models using conditional flow
matching. The decoded distribution enables two downstream uses without
modifying the policy: online predictive tuning of a high-level controller
around the fixed low-level policy, and robustness analysis under specified
disturbances. Under perturbed actuator dynamics, decoded-model predictive
tuning reduces position tracking RMSE by $23\%$ and heading RMSE by $45\%$
relative to fixed gains. Under Gaussian force disturbances, decoded-model
ensembles closely predict the lateral tracking-error evolution. Together,
these results show that control latents can be converted into physical model
ensembles for tuning, robustness analysis, and diagnosis of frozen adaptive
policies.
    \end{abstract}



    \section{Introduction}

Latent-based adaptive control has enabled robots to operate across environments where dynamics vary. The standard paradigm, popularized by frameworks such as~\cite{RMA2021kumar, UP-OSI2017yu, pearl2019rakelly}, infers a compact latent representation from recent sensorimotor history, and a control policy conditioned on this latent adapts online to shifting conditions. By learning a task-sufficient representation~\cite{alemi2017deep} rather than identifying the full system, this pipeline has been applied to legged locomotion~\cite{RMA2021kumar, kumar2022adapting}, dexterous in-hand manipulation~\cite{qi2023hand}, and quadrotor flight~\cite{zhang2023xadapprev, zhang2025xadap}. However, a conceptual gap remains: these latent representations lack grounded physical meaning. While they capture the information necessary for control, the specific physical structures or parameters they encode remain opaque. This opacity complicates safety certification and post-hoc diagnosis of the learned controller. Prior work evaluates these latents primarily through downstream task success, leaving the physical content of these latents unexamined.

A primary reason for this opacity is that the inverse mapping from observed behavior to physical parameters is ill-posed~\cite{Ljung1999}. In robotics, the relationship between behavior and dynamics is often non-injective: many physically distinct systems can exhibit indistinguishable behavior under the same trajectories. For example, a heavier quadcopter can exhibit the same closed-loop motion as a lighter vehicle with weaker actuation. We hypothesize that latent-based methods do not learn unique physical constants, but rather encode equivalence classes of plausible dynamics models. Consequently, approaches in this spirit that regress each latent to a single set of ground-truth parameters (e.g.,~\cite{UP-OSI2017yu}) are limited, because such one-to-one mappings cannot represent this physical non-identifiability.

In this work, we propose to interpret latent representations by explicitly modeling the one-to-many structure of the latent-to-parameter map. Rather than regressing each latent to a single point estimate of physical parameters, we use conditional flow matching~\cite{lipman2022flow} to decode each latent into a continuous distribution over the physically distinct models consistent with it. We instantiate the framework on quadrotor systems, where the decoded distribution supports two downstream uses without modifying the policy: online predictive tuning of a high-level controller around the fixed policy, and robustness analysis under Gaussian force disturbances. We then evaluate whether the decoded parameters explain the policy's adaptation by comparing motor commands from decoded parametric controllers against those from the latent-conditioned policy under identical hardware state histories.
\begin{figure}
    \centering
    \includegraphics[width=\linewidth]{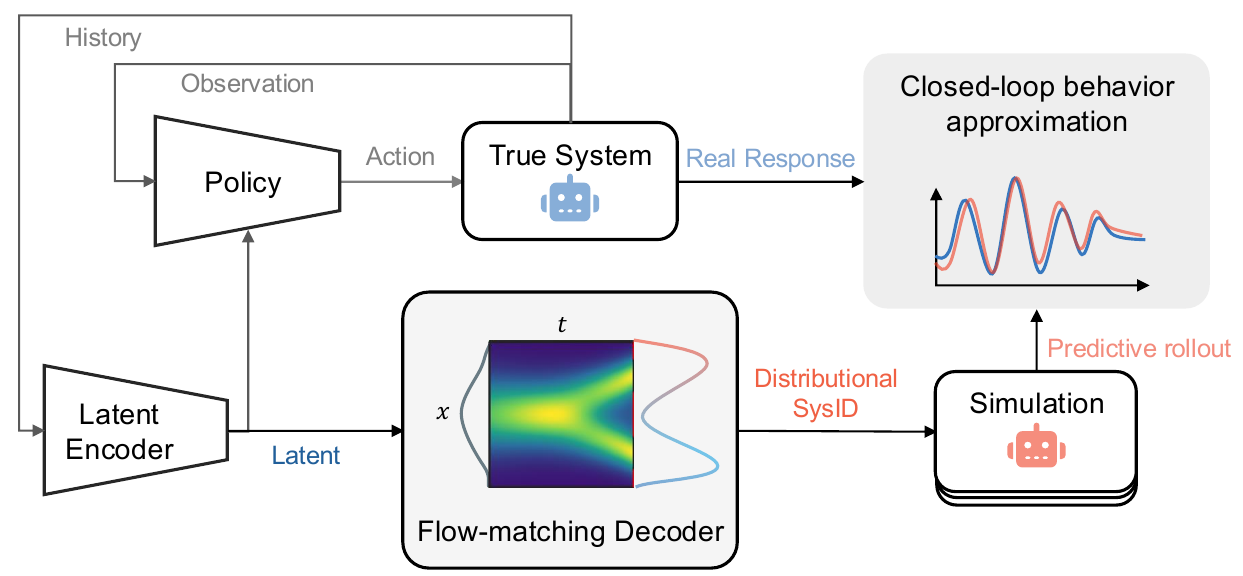}
    \caption{Overview of the proposed interpretation framework. Observation history is encoded into a compact latent, which a conditional flow-matching decoder then maps to a distribution of physically consistent system models (distributional system identification). Each decoded model is interpretable and approximates the true system's closed-loop behavior under the deployed policy: they are run in parallel simulations (parametrized by the decoded models) and predictive rollouts are compared against real responses, enabling downstream tasks such as controller tuning and robustness analysis without modifying the policy.}
    \label{fig:face}
\end{figure}

Conceptually, our approach is inspired by model reference adaptive control (MRAC)~\cite{ioannou2010robust}, where a nominal reference model defines the desired closed-loop behavior and adaptation drives the controller to match it. Rather than tune a controller toward a fixed reference, we recover a family of reference models from the latent itself. The decoded models can run as parallel simulation environments that approximate the learned policy's closed-loop behavior, supporting analysis without further hardware experiments.

    \section{Related Work} \label{sec:related_work}

Latent representations provide compact encodings of high-dimensional observations, capturing task-relevant structure for downstream planning and control~\cite{kingma2013auto, alemi2017deep}. In robotics, they are widely used to enable adaptive control. Frameworks such as~\cite{RMA2021kumar, lee2020learning} infer a latent vector from recent proprioceptive history and condition a fixed policy on this embedding for fast online adaptation. More recent variants use world-model rollouts as implicit reference trajectories for rapid motor adaptation, separating long-horizon reward optimization from fast latent-space corrections~\cite{brito2025reflexive}. Meta-reinforcement learning methods~\cite{pearl2019rakelly, duan2016rl2} treat latents as task beliefs to support few-shot generalization, while world-model approaches~\cite{hafner2019dream, hafner2023mastering} learn latent dynamics for planning in compact state spaces. In these works the latent serves as a control-sufficient signal, and its physical content is not examined directly.

Another line of work examines what learned representations encode about the physical world. Physically Interpretable World Models~\cite{mao2024piwm} align latent states with known physical variables through weak distribution-based supervision and constrain their evolution through partially known dynamics. Phys2Real~\cite{wang2025phys2real} fuses vision-language-model priors with online adaptation to condition manipulation policies directly on interpretable physical parameters such as center of mass, using ensemble-based uncertainty quantification. These methods shape the representation during training so that latent dimensions correspond to physical quantities.

Classical adaptive control takes a complementary, model-based perspective, maintaining online estimates of parameters or matched uncertainties to guarantee desired closed-loop behavior. Indirect Model Reference Adaptive Control (MRAC)~\cite{ioannou2010robust} and $\mathcal{L}_1$ adaptive control~\cite{hovakimyan2010l1} follow this line. System identification methods recover dynamics models from input/output data; traditional techniques such as prediction-error minimization~\cite{Ljung1999} and subspace identification~\cite{van2012subspace} return point estimates of system parameters and typically assume identifiability. Structural identifiability theory~\cite{bellman1970structural} shows, however, that inverse mappings from behavior to physical parameters are not always unique: multiple distinct systems can produce indistinguishable responses under limited excitation, so a single best-fit estimate does not capture the ambiguity in the solution set.

Probabilistic system identification~\cite{beck2010bayesian, peterka1981bayesian} represents parameters as distributions rather than single estimates. Many of these methods return a uni-modal posterior that concentrates around a single model: this captures estimation uncertainty, but not physical non-identifiability, where a whole set of distinct models is simultaneously consistent with the observed behavior. Representing that structure requires a flexible distribution able to cover the full solution set, whose shape is not known a priori and may be broad or even multi-modal. In robotics, simulation-based inference approximates posteriors over simulator parameters when the likelihood is intractable. BayesSim~\cite{ramos2019bayessim} estimates them with mixture-density networks and uses them to drive adaptive domain randomization, and later work leverages differentiable simulation to obtain gradient-based posterior particles~\cite{heiden2022probabilistic}. The central change in viewpoint we adopt is thus from point estimation to distributional estimation that represents the full set of physically distinct models compatible with a given latent, rather than a single best-fit model with error bars.

Diffusion and flow-based generative models~\cite{ho2020denoising, lipman2022flow, tong2023improving} model complex, multi-modal distributions on nonlinear manifolds. Recent work applies such models to parameter estimation in nonlinear dynamical systems, using joint conditioning on multiple observations to mitigate non-identifiability in the inverse mapping~\cite{zhu2024jcdi}. Their intersection, in which a generative model decodes a conditioning signal into a distribution of physical parameters consistent with it, is the setting of this paper.

\section{Methodology}

\begin{figure}
    \centering
    \includegraphics[width=\linewidth]{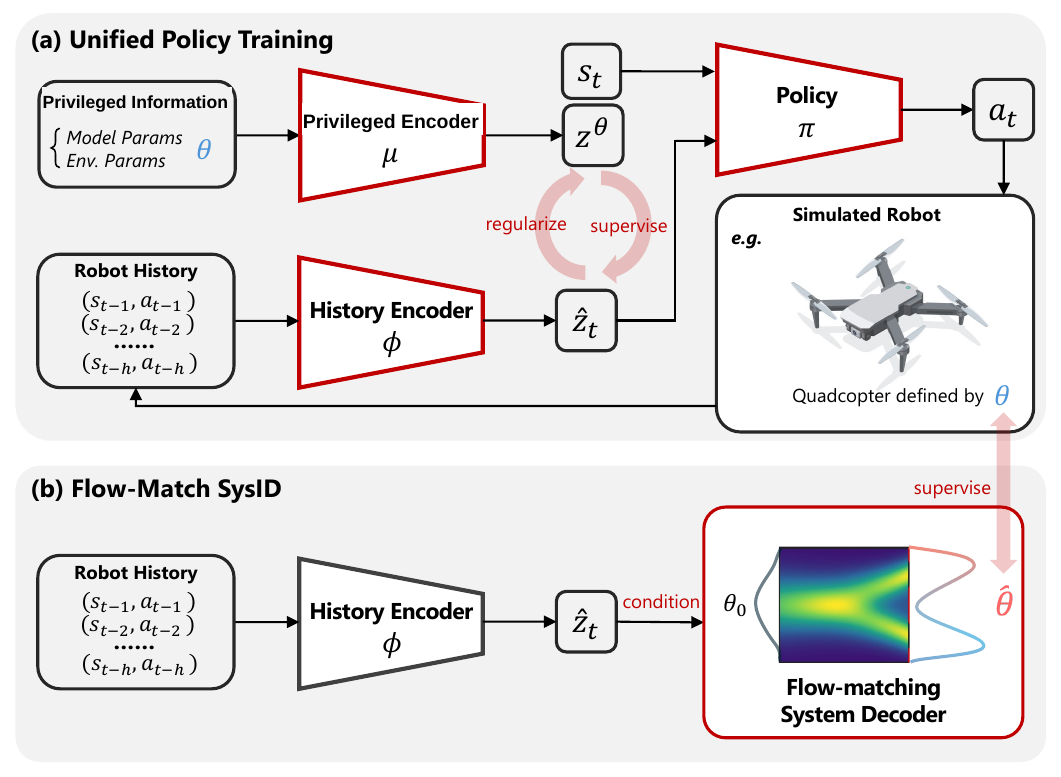}
    \caption{Overview of the training pipeline. (a) Unified training aligns the privileged latent with the history-based latent so the policy can adapt across randomized dynamics. (b) Conditional flow matching decodes the latent into a distribution of dynamically consistent physical models.}
    \label{fig:overview}
\end{figure}

Our framework consists of two components:
(1) learning a latent-based adaptive controller, and
(2) decoding the latent representation into a distribution of dynamically consistent models via conditional flow matching. An overview is shown in Figure~\ref{fig:overview}.

\subsection{Unified Latent-Based Adaptive Control} \label{sec:unify}

We consider a dynamical system with states $\bm{s}_t \in \mathcal{S}$,
actions $\bm{a}_t \in \mathcal{A}$, and physical parameters $\theta \in \Theta$.
The parameters $\theta$ are unavailable at deployment.
Our goal is to learn a control policy that adapts to environment-dependent
dynamics without direct access to $\theta$.
We learn a policy $\pi(\bm{a}_t \,|\, \bm{s}_t, \bm{z}_t)$ conditioned on a latent variable
$\bm{z}_t \in \mathbb{R}^{d_z}$.
During training, the latent is produced by two encoders:
\begin{equation}
\bm{z}^\theta = \mu(\theta), 
\qquad
\hat{\bm{z}}_t = \phi(\tau_{t-h:t}),
\end{equation}
where $\mu$ maps physical parameters to a privileged latent and
$\phi$ estimates a latent from recent sensorimotor history
$\tau_{t-h:t} = \{\bm{s}_{t-h:t}, \bm{a}_{t-h:t}\}$.
At deployment, only $\hat{\bm{z}}_t$ is available and the policy operates as
\begin{equation}
\bm{a}_t \sim \pi(\bm{a}_t \,|\, \bm{s}_t, \hat{\bm{z}}_t).
\end{equation}

Traditional approaches~\cite{lee2020learning,RMA2021kumar} train $\mu$ and $\pi$ first using privileged
information and subsequently train $\phi$ to imitate the teacher.
This two-stage scheme leaves a gap between $\bm{z}^\theta$
and $\hat{\bm{z}}_t$, because $\mu$ can exploit information
not recoverable from sensorimotor history. 

We instead optimize $\pi$, $\mu$, and $\phi$ jointly, inspired by~\cite{fu2023deep}.
This unified training produces a latent space that is
consistent across privileged and history-based representations, reducing the mismatch between
$\bm{z}^\theta$ and $\hat{\bm{z}}_t$.
The latent must also serve as a stable conditioning variable for the
subsequent system identification stage: a representation trained separately from the
policy, or one that collapses, no longer encodes the physical variation
needed for decoding.
Joint optimization avoids both failure modes by keeping the latent
control-sufficient and recoverable from history.
During training, the policy is conditioned on
\begin{equation}
\bm{z}_t =
\begin{cases}
\bm{z}^\theta, & \text{with probability } 1-p(t), \\
\hat{\bm{z}}_t, & \text{with probability } p(t),
\end{cases}
\end{equation}
where $p(t)$ increases from $0$ to $p_{\max}$ after an initial warmup
period. This scheduled substitution gradually couples the adaptation
module to the policy.

The reinforcement learning objective is
\begin{equation}
J(\theta_\pi, \theta_\mu)
=
\mathbb{E}_{\pi}\!\left[\sum_{t=0}^{T} \gamma^t r_t \right].
\end{equation}
Naively optimizing the unified objective may lead to a degenerate
solution in which the latent representation collapses to a constant
vector, allowing the policy to ignore environment variation.
To promote observability and prevent collapse, we augment the objective with alignment and regularization terms, yielding the joint loss
\begin{equation}
\begin{aligned}
\mathcal{L}(\theta_\pi, \theta_\mu, \theta_\phi)
=
&- J(\theta_\pi, \theta_\mu)
+ \lambda_1 \| \bm{z}^\theta - \mathrm{sg}[\hat{\bm{z}}_t] \|_2^2 \\
&+ \lambda_2 \| \mathrm{sg}[\bm{z}^\theta] - \hat{\bm{z}}_t \|_2^2
+ \mathcal{L}_{\text{aux}} ,
\end{aligned}
\end{equation}
where $\mathrm{sg}[\cdot]$ denotes the stop-gradient operator.
The first alignment term constrains the privileged latent
$\bm{z}^\theta$ to remain predictable from sensorimotor history,
while the second trains the adaptation module to track
the privileged encoder online.
The auxiliary term $\mathcal{L}_{\text{aux}}$ aggregates additional
stabilization objectives, including a variance regularization on
$\bm{z}^\theta$ to prevent representational collapse and a reconstruction
loss that maps $\bm{z}^\theta$ back to the physical parameters $\theta$
through a decoder network.
After convergence, unified training enforces consistency between the
privileged latent $\bm{z}^\theta = \mu(\theta)$ and the history-based
estimate $\hat{\bm{z}}_t$, so both live on a shared latent manifold.
Because only $\hat{\bm{z}}_t$ is observable at deployment, the system
identification stage described next conditions on $\hat{\bm{z}}_t$ throughout training
and inference.

\subsection{Generative System Identification via Conditional Flow Matching}

We model the conditional distribution of physical parameters given the operational latent,
\begin{equation}
\theta \sim p(\theta  \,|\, \hat{\bm{z}}_t),
\end{equation}
using Conditional Flow Matching (CFM)~\cite{lipman2022flow}. CFM defines a probability path
$p_\tau(\theta \,|\, \hat{\bm{z}}_t)$ that transports a simple base distribution
$p_0(\theta)=\mathcal{N}(0,I)$ to the empirical parameter distribution
conditioned on the latent.
A neural network $v_\psi(\theta_\tau, \tau, \hat{\bm{z}}_t)$ approximates the conditional
vector field governing this transport.

We adopt a Gaussian-smoothed linear interpolation path
between $\theta_0 \sim \mathcal{N}(0,I)$ and $\theta_1$ sampled from
the dataset of physical parameters:
\begin{equation}
\theta_\tau = (1 - \tau)\,\theta_0 + \tau\,\theta_1 + \sigma\bm{\varepsilon},
\quad
\bm{\varepsilon} \sim \mathcal{N}(0,I),
\quad \tau \sim \mathcal{U}[0,\tau_{\max}],
\end{equation}
where $\sigma > 0$ is a small smoothing parameter.
The perturbation prevents the path from becoming degenerate and
smooths the marginal velocity field. Truncating the flow time to
$[0, \tau_{\max}]$ with $\tau_{\max} < 1$ avoids ill-conditioned
regression near $\tau = 1$, where the smoothing noise
$\sigma\bm{\varepsilon}$ becomes comparable to the residual signal
$(1-\tau)(\theta_1 - \theta_0)$.
The target conditional vector field is
\begin{equation}
u_\tau(\theta_0,\theta_1) = \theta_1 - \theta_0,
\end{equation}
which is independent of the noise realization $\bm{\varepsilon}$.
The flow-matching objective is therefore
\begin{equation}
\mathcal{L}_{\text{CFM}}(\psi)
=
\mathbb{E}_{\tau,\, \theta_0,\, \theta_1,\, \bm{\varepsilon}}
\left\|
v_\psi(\theta_\tau, \tau, \hat{\bm{z}}_t)
-
(\theta_1 - \theta_0)
\right\|^2,
\end{equation}
where $\hat{\bm{z}}_t$ is the operational latent associated with $\theta_1$
through the unified training process.

At inference, given a latent $\hat{\bm{z}}_t$ obtained from
sensorimotor history, a sample of physical parameters $\hat{\theta}$
is generated by integrating the learned vector field from
$\theta_0 \sim \mathcal{N}(0, I)$:
\begin{equation}
\frac{d\theta}{d\tau} = v_\psi(\theta_\tau, \tau, \hat{\bm{z}}_t),
\quad \theta_{\tau=0} \sim \mathcal{N}(0, I),
\end{equation}
from $\tau=0$ to $\tau=1$ using a fixed-step Euler solver.
The resulting $\hat{\theta}$ is a sample from the learned
conditional distribution $p(\theta \,|\, \hat{\bm{z}}_t)$, giving a
dynamically consistent estimate of the system parameters.

    \section{Implementation}

While the proposed framework applies to general parametric dynamical systems, we instantiate it on quadrotor platforms following the cross-platform adaptation setting of~\cite{zhang2025xadap}. This instantiation is motivated by two considerations. First, quadrotors admit accurate first-principles modeling and a standard parametric controller, which lets decoded parameters be evaluated against policy actions. Second, cross-platform adaptation spans orders-of-magnitude variation in mass, inertia, and motor constants, providing a broad and well-characterized distribution of systems from which to decode.

\subsection{Training Setup}

The base adaptive policy $\pi$, privileged encoder $\mu$, and adaptation module $\phi$ follow the architecture in~\cite{zhang2025xadap} and produce a latent representation $\bm{z} \in \mathbb{R}^{8}$. The policy acts as a platform-agnostic inner-loop controller: it receives IMU measurements (angular velocity and linear acceleration) together with high-level commands (desired mass-normalized collective thrust ${a}_{\mathrm{cmd}}$ and desired angular velocity $\bm{\omega}_{\mathrm{cmd}}$), and outputs normalized motor speed commands $\bm{\Omega}_{\mathrm{norm}} = \bm{\Omega}_{\mathrm{cmd}}/\Omega_{\max} \in [0,1]^4$, where $\Omega_{\max}$ is the maximum rotor speed of the platform. The body-frame and motor-indexing conventions used throughout are shown in Figure~\ref{fig:drone}. Unified training is performed in simulation under extensive domain randomization over physical parameters, covering cross-platform quadrotor variations spanning orders of magnitude as defined in~\cite{zhang2025xadap}. The policy is trained with scheduled substitution between $\bm{z}^\theta$ and $\hat{\bm{z}}_t$ as described in Section~\ref{sec:unify}, ensuring alignment between privileged and history-based representations.

After convergence, we collect a dataset of $20,000$ parameter and latent pairs $(\theta, \hat{\bm{z}})$ by sampling physical parameters from the training distribution and recording the corresponding operational latent representation produced by the adaptation module. This dataset is used to train the conditional flow matching model. The flow-matching velocity field $v_{\psi}$ is implemented as a 4-layer residual network with hidden dimension $512$ and SiLU activations~\cite{elfwing2018sigmoid}. The noisy parameter vector $\theta_t$ is embedded by a two-layer MLP and the diffusion time $t$ is encoded via a sinusoidal positional embedding~\cite{vaswani2017attention} followed by a two-layer MLP; the two embeddings are concatenated and linearly projected to the hidden dimension. The latent condition $\hat{\bm{z}}$ is mapped to the same hidden dimension by a three-layer MLP, and at each residual block its output modulates the intermediate features through FiLM~\cite{perez2018film} scale-and-shift parameters. The network contains approximately $5$M parameters. We train using CFM~\cite{lipman2022flow} with $\sigma=0.05$, the AdamW optimizer~\cite{adamw} with learning rate $5{\times}10^{-4}$, weight decay $10^{-5}$, gradient clipping at $1.0$, and cosine annealing over $100$ epochs with a batch size of $256$. The flow time $t$ is sampled uniformly from $[0,\,0.95]$ during training. At inference, the learned vector field is integrated using Euler's method with $50$ fixed steps.

\subsection{Parameter Space}
The parameter vector $\theta \in \mathbb{R}^{19}$ is defined as Equation~\eqref{eq:param}, where $m$ denotes the mass, $l$ the arm length, $k_t$ the thrust coefficient, and $k_\tau$ the torque coefficient. The inertia matrix is $\bm J=\mathrm{diag}(J_{xx},J_{yy},J_{zz})$, the linear drag matrix is $\bm D\in\mathbb{R}^{3\times3}$ (diagonal), $\Omega_{\max}$ is the maximum rotor speed, $\bm \eta\in\mathbb{R}^4$ are motor effectiveness factors, $\tau_m$ is the motor time constant, and $\bm \tau_{\mathrm{ext}}\in\mathbb{R}^3$ represents external torque disturbances. The randomization range follows~\cite{zhang2025xadap} and spans large variations to ensure that the latent representation captures substantially different flight regimes.
\begin{equation}
\theta =
\big(
m,\;
l,\;
k_t,\;
k_\tau,\;
J_{xx}, J_{yy}, J_{zz},\;
\bm D,\;
\Omega_{\max},\;
\bm \eta,\;
\tau_m,\;
\bm \tau_{\mathrm{ext}}
\big)
\label{eq:param}
\end{equation}

\subsection{Experimental Setup}
\label{sec:setup}
\paragraph{Simulation Environment}

All unified training is conducted in Flightmare~\cite{song2021flightmare}, which provides quadrotor dynamics with parameter randomization over the defined physical space. The trained policy, adaptation module, and generative model are evaluated in simulation before hardware deployment.

\paragraph{Hardware Platform} 
\begin{figure}
  \centering
  \includegraphics[width=0.8\linewidth]{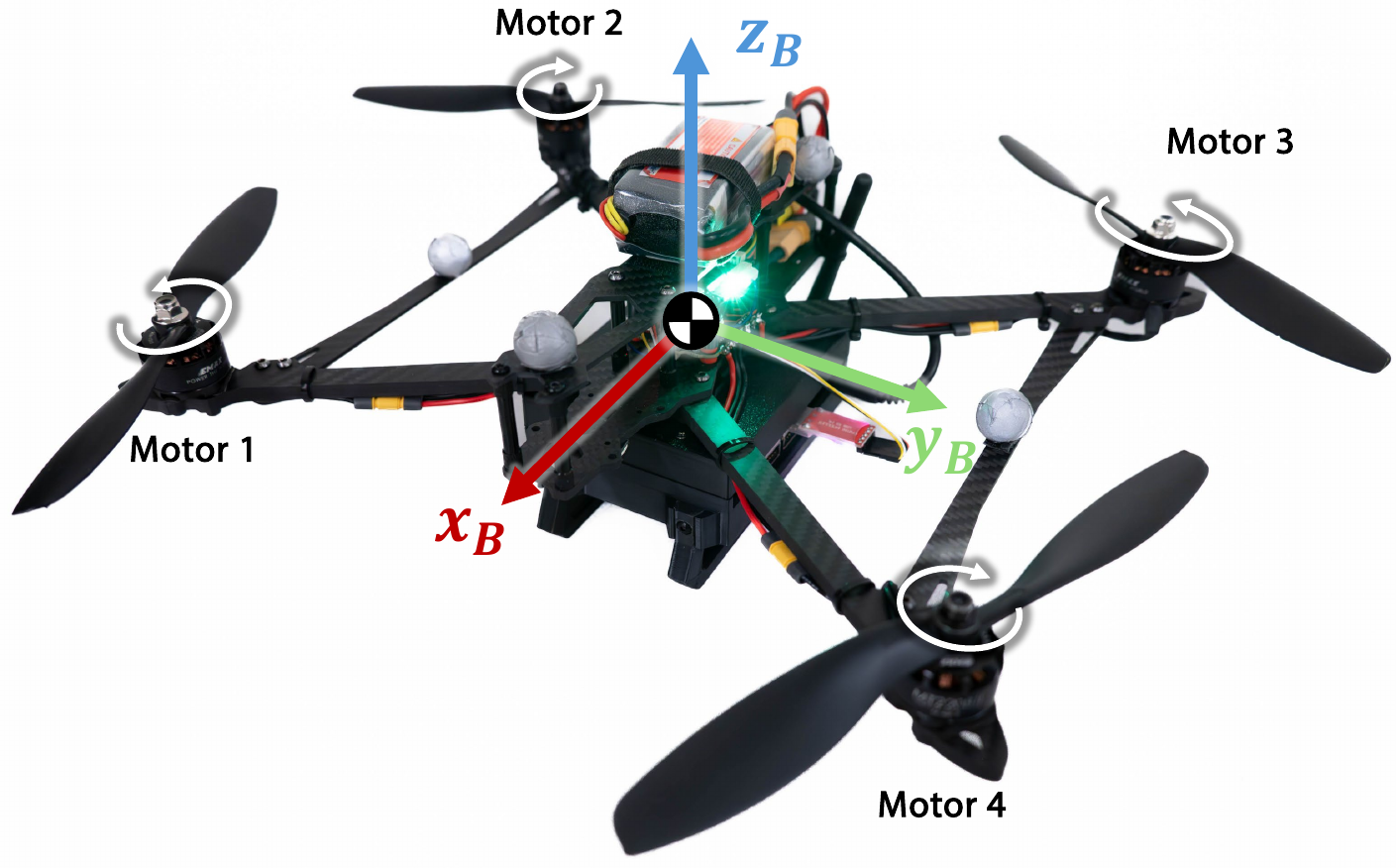}
  \caption{Quadrotor convention used in this work. The body-fixed reference frame consists of the orthogonal basis $\bm{x}_B,\bm{y}_B,\bm{z}_B$. Four motor-propeller pairs are arranged symmetrically about the vehicle center of mass. Motors 1 and 3 rotate in the opposite direction from motors 2 and 4. The nominal hardware parameters, measured directly from the vehicle and its components, are listed in Table~\ref{tab:quad-params}.}
  \label{fig:drone}
\end{figure}

\begin{table}
  \centering
  \caption{Quadcopter Parameters}
   \begin{tabular}{c|c|c}
    \toprule
    Parameter                   & Value                         & Unit\\
    \midrule
    $m$                         & $0.925$                       & $\mathrm{kg}$\\
    $l$                         & $0.184$                       & $\mathrm{m}$\\
    $\bm J$                     & $\mathrm{diag}(8.5,7.5,15.5)$ & $\mathrm{g\cdot m^2}$\\
    $\bm D$                     & $\mathrm{diag}(0.20, 0.20, 0.85)$ & $\mathrm{kg/s}$\\
    $k_t$                       & $7.64e^{-6}$                  & $\mathrm{N/(rad/s)^2}$\\
    $k_\tau$                    & $1.07e^{-7}$                  & $\mathrm{N\cdot m/(rad/s)^2}$\\
    $\tau_{\mathrm{m}}$         & $0.03$                        & $\mathrm{s}$\\
    $\Omega_{\max}$             & $1360$                        & $\mathrm{rad/s}$\\
    \bottomrule
  \end{tabular}
  \label{tab:quad-params}
\end{table}

Figure~\ref{fig:drone} shows the quadrotor used for real-world experiments. Its nominal parameters, listed in Table~\ref{tab:quad-params}, are obtained independently of the latent decoder by direct measurement and bench characterization: mass, arm length, and geometry are measured directly from the vehicle, and the thrust and torque coefficients are obtained from static motor bench tests. The platform is equipped with four 2208 1500kV brushless motors and 8045-2 propellers. The Pixracer R15 flight control unit provides IMU measurements, including linear acceleration and body rates. The adaptation module and policy network are deployed on an NVIDIA Jetson Orin Nano and execute at $500\,\mathrm{Hz}$. Control commands are transmitted from the onboard computer to the flight controller via a USB-TTL link and subsequently executed by the electronic speed controllers.

    \section{Decoded Models for Control and Robustness Analysis}
\label{sec:application}

We examine two representative uses of the decoded models around the fixed latent-conditioned policy: online predictive tuning of a high-level controller and robustness analysis under specified disturbance distributions.

\subsection{Predictive Control on the Decoded Models}

The low-level policy tracks high-level commands, including desired mass-normalized collective thrust and desired body rates, and remains frozen after training. A high-level controller converts the reference trajectory into these commands. Here, we use the decoded models to tune this high-level controller online, while keeping the neural policy unchanged.

The \textit{CFM-tuned} controller starts from naive gains and updates the high-level controller at $10\,\mathrm{Hz}$, matching the CFM inference rate used in the action-replay experiments. At each update, CFM samples decoded models from the current latent, initializes parallel predictive simulations from the current state, and rolls out the frozen policy under candidate high-level controller settings. A cross-entropy method (CEM) optimizer~\cite{deboer2005cem} then selects the setting with the lowest short-horizon cost, which penalizes position tracking error, motor saturation, and rollout failure. In this use case, the decoded models provide local predictive rollouts for online command tuning.

\paragraph{Nominal Predictive Control}
As a nominal check, we evaluate predictive tuning for $5\,\mathrm{s}$ on $100$ randomized quadrotor environments, each tracking online random trajectories~\cite{mellinger2011minimum}. We compare three high-level controllers: \textit{Naive}, which uses uninformed default gains; \textit{Baseline}, a hand-tuned reference controller; and \textit{CFM-tuned}, which starts from the naive gains and updates them online using decoded-model rollouts. Table~\ref{tab:predictive_nominal} reports position RMSE, maximum position error, and heading RMSE.

\begin{table}[t]
    \centering
    \caption{Predictive high-level controller tuning over $100$ online random trajectories with randomized quadrotor dynamics. Values are mean $\pm$ standard deviation across rollouts.}
    \label{tab:predictive_nominal}
    \small
    \setlength{\tabcolsep}{3pt}
    \resizebox{\linewidth}{!}{%
    \begin{tabular}{@{}lccc@{}}
    \toprule
    \textbf{Metric} & \textbf{Naive} & \textbf{Baseline} & \textbf{CFM-tuned} \\
    \midrule
    \multirow{2}{*}{\begin{tabular}[c]{@{}l@{}}Pos. RMSE\\(m)\end{tabular}}
        & \multirow{2}{*}{$0.330 \pm 0.147$}
        & \multirow{2}{*}{$\mathbf{0.198 \pm 0.103}$}
        & \multirow{2}{*}{$0.218 \pm 0.102$} \\
        & & & \\
    \multirow{2}{*}{\begin{tabular}[c]{@{}l@{}}Max Pos. Err.\\(m)\end{tabular}}
        & \multirow{2}{*}{$0.624 \pm 0.291$}
        & \multirow{2}{*}{$\mathbf{0.470 \pm 0.298}$}
        & \multirow{2}{*}{$0.510 \pm 0.268$} \\
        & & & \\
    \multirow{2}{*}{\begin{tabular}[c]{@{}l@{}}Heading RMSE\\(deg)\end{tabular}}
        & \multirow{2}{*}{$\mathbf{2.676 \pm 1.883}$}
        & \multirow{2}{*}{$4.696 \pm 2.724$}
        & \multirow{2}{*}{$4.894 \pm 3.368$} \\
        & & & \\
    \bottomrule
    \end{tabular}
    }
\end{table}

In the nominal setting, \textit{CFM-tuned} improves over \textit{Naive}, but does not surpass \textit{Baseline}. It reduces position RMSE by $34\%$ and maximum position error by $18\%$ relative to \textit{Naive}, closing most of the gap to \textit{Baseline} using only decoded-model rollouts. Even when started from uninformed gains, the decoded models are therefore accurate enough to support a competitive predictive tuner. \textit{Baseline}, in contrast, is iteratively retuned offline against the nominal system through trial and error.

\paragraph{Predictive Control with Slower Actuator Dynamics}
Actuator latency and state-transition delay are well-known failure modes for learning-based control policies, where they measurably degrade closed-loop performance~\cite{sandha2021sim2real, tan2018simtoreal}. Motivated by this, we deliberately stress the same fixed policy with slower actuators by widening the motor first-order time constant from the nominal $\tau_m \in [0.02,\,0.04]\,\mathrm{s}$ to $\tau_m \in [0.05,\,0.10]\,\mathrm{s}$. The slower actuator response lengthens the effective input delay and pushes the frozen policy toward saturation under aggressive high-level commands, putting it in a regime where the policy alone is expected to track poorly. We reuse the same CFM-based tuner without retraining the policy or changing the cost, which tests whether the CFM can decode the slower actuators from the latent and detune the high-level controller before saturation occurs.

\begin{table}[t]
    \centering
    \caption{Predictive tuning with increased motor response time constants. Values are mean $\pm$ standard deviation over $100$ online random trajectories.}
    \label{tab:predictive_latency}
    \small
    \setlength{\tabcolsep}{3pt}
    \resizebox{\linewidth}{!}{%
    \begin{tabular}{@{}lccc@{}}
    \toprule
    \textbf{Metric} & \textbf{Baseline} & \begin{tabular}[c]{@{}c@{}}\textbf{Nominal}\\\textbf{+CEM}\end{tabular} & \textbf{CFM-tuned} \\
    \midrule
    \multirow{2}{*}{\begin{tabular}[c]{@{}l@{}}Pos. RMSE\\(m)\end{tabular}}
        & \multirow{2}{*}{$0.705 \pm 1.256$}
        & \multirow{2}{*}{$0.584 \pm 0.725$}
        & \multirow{2}{*}{$\mathbf{0.546 \pm 0.560}$} \\
        & & & \\
    \multirow{2}{*}{\begin{tabular}[c]{@{}l@{}}Max Pos. Err.\\(m)\end{tabular}}
        & \multirow{2}{*}{$1.671 \pm 3.144$}
        & \multirow{2}{*}{$1.332 \pm 2.078$}
        & \multirow{2}{*}{$\mathbf{1.155 \pm 1.363}$} \\
        & & & \\
    \multirow{2}{*}{\begin{tabular}[c]{@{}l@{}}Heading RMSE\\(deg)\end{tabular}}
        & \multirow{2}{*}{$14.251 \pm 15.581$}
        & \multirow{2}{*}{$13.807 \pm 13.441$}
        & \multirow{2}{*}{$\mathbf{7.853 \pm 6.581}$} \\
        & & & \\
    \bottomrule
    \end{tabular}
    }
\end{table}



\textit{CFM-tuned} achieves the best performance under the stressed actuator dynamics, reducing position RMSE by $23\%$, maximum position error by $31\%$, and heading RMSE by $45\%$ over \textit{Baseline}. To separate the decoded ensemble from the CEM optimizer itself, the \textit{Nominal+CEM} ablation runs the same CEM tuner over the nominal system with $\pm 20\%$ parameter perturbations. It improves over \textit{Baseline} but remains worse than \textit{CFM-tuned} on every metric, so the gains come from the decoded models rather than from the optimizer alone. We note that \textit{CFM-tuned} reaches this performance zero-shot and without any nominal-system information, with both the policy and the decoder driven solely by the observation history. The \textit{Nominal+CEM} ablation, by comparison, has access to the nominal parameters under a bounded $\pm 20\%$ uncertainty set.

\paragraph{Hardware Gain Tuning}

We finally deploy the same predictive tuner on hardware for 3D lemniscate tracking, using slow, medium, and fast references that complete one cycle in $15\,\mathrm{s}$, $10\,\mathrm{s}$, and $7.5\,\mathrm{s}$, respectively. These references excite all three translational axes and test whether the tuner transfers from simulation to hardware. \textit{Baseline} is the hand-tuned controller used in the nominal experiments, while \textit{CFM-tuned} starts from the same naive gains as above and updates them online from decoded-model rollouts.

\begin{figure}[t]
    \centering
    \includegraphics[width=\linewidth]{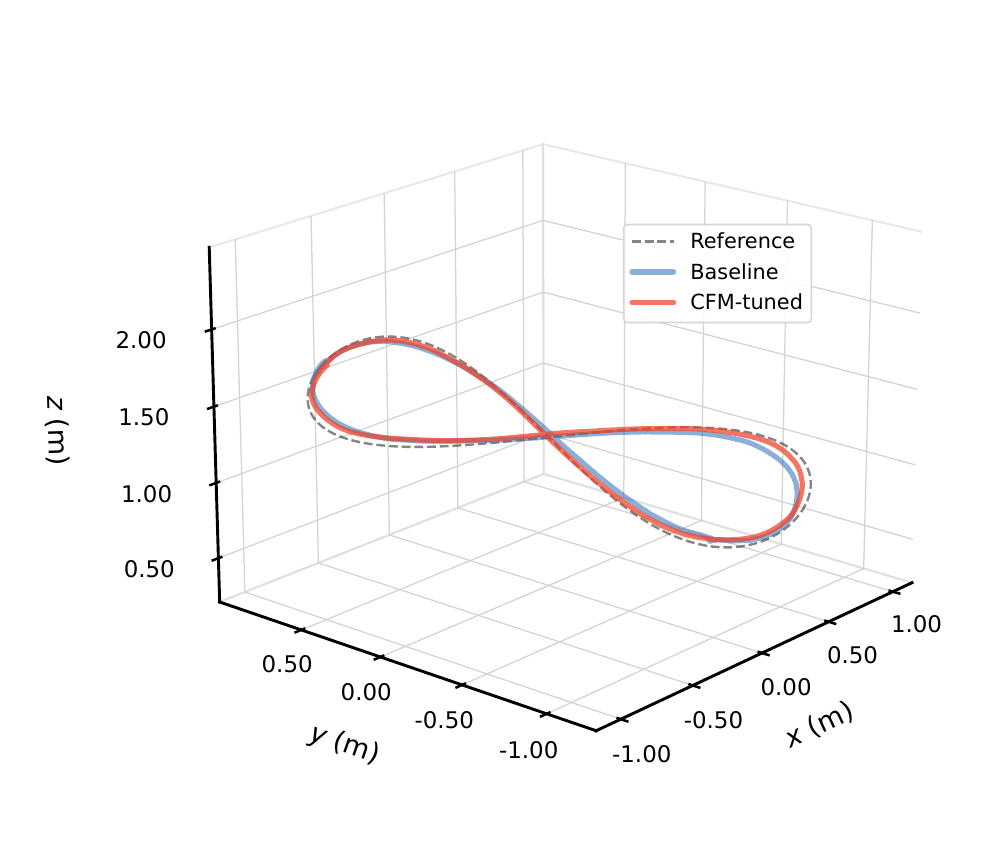}
    \caption{Hardware tracking of the 3D lemniscate at medium reference speed. \textit{CFM-tuned} achieves tracking comparable to \textit{Baseline} while reducing the spread of the position error, consistent with Table~\ref{tab:hardware_gain_tuning}.}
    \label{fig:tracking_3d_mid}
\end{figure}

Table~\ref{tab:hardware_gain_tuning} shows that \textit{CFM-tuned} tracks comparably to \textit{Baseline} across all three hardware speeds, with mean position errors differing by only a few centimeters. \textit{CFM-tuned} also lowers both the peak position error and the position-error standard deviation at every speed, suggesting that it picks less extreme gains than \textit{Baseline} while matching average accuracy. Figure~\ref{fig:tracking_3d_mid} visualizes the medium-speed trial, where \textit{CFM-tuned} follows the 3D reference with reduced tracking spread.

\begin{table}[t]
    \centering
    \caption{Hardware gain tuning for 3D lemniscate tracking at slow, medium, and fast reference speeds (three complete cycles each). Position and heading columns report RMSE $\pm$ std.\ of per-sample error; max position error is the peak over the window. Bold marks the lower value for each speed and metric.}
    \label{tab:hardware_gain_tuning}
    \small
    \setlength{\tabcolsep}{4.5pt}
    {\renewcommand{\arraystretch}{1.4}%
    \begin{tabular}{@{}llccc@{}}
    \toprule
    \textbf{Speed} & \textbf{Ctrl.} &
    \begin{tabular}[c]{@{}c@{}}Pos.\ RMSE\\(m)\end{tabular} &
    \begin{tabular}[c]{@{}c@{}}Max pos.\\err.\ (m)\end{tabular} &
    \begin{tabular}[c]{@{}c@{}}Heading RMSE\\(deg)\end{tabular} \\
    \midrule
    \multirow{2}{*}{Slow}
        & \textit{Baseline}
        & $0.046 \pm 0.012$
        & $0.079$
        & $\mathbf{1.197 \pm 0.177}$ \\
        & \textit{CFM-tuned}
        & $\mathbf{0.025 \pm 0.009}$
        & $\mathbf{0.048}$
        & $1.303 \pm 0.302$ \\
    \addlinespace[0.4em]
    \midrule
    \multirow{2}{*}{Medium}
        & \textit{Baseline}
        & $0.060 \pm 0.015$
        & $0.090$
        & $1.316 \pm 0.291$ \\
        & \textit{CFM-tuned}
        & $\mathbf{0.038 \pm 0.010}$
        & $\mathbf{0.062}$
        & $\mathbf{1.258 \pm 0.268}$ \\
    \addlinespace[0.4em]
    \midrule
    \multirow{2}{*}{Fast}
        & \textit{Baseline}
        & $\mathbf{0.110 \pm 0.045}$
        & $0.252$
        & $\mathbf{1.354 \pm 0.341}$ \\
        & \textit{CFM-tuned}
        & $0.129 \pm 0.035$
        & $\mathbf{0.187}$
        & $1.386 \pm 0.365$ \\
    \bottomrule
    \end{tabular}%
    }
\end{table}

\subsection{Robustness Analysis under Disturbances}
\label{sec:uncertainty-quant}

We next use the decoded models to estimate robustness and uncertainty of the fixed policy under external disturbances. For each latent, CFM gives a distribution of physically plausible quadrotor models. Rolling out this model distribution under the same policy and disturbance gives an uncertainty envelope for the closed-loop response.

\subsubsection{Individual Step-Disturbance Analysis}

\begin{figure}[t]
    \centering
    \includegraphics[width=\linewidth]{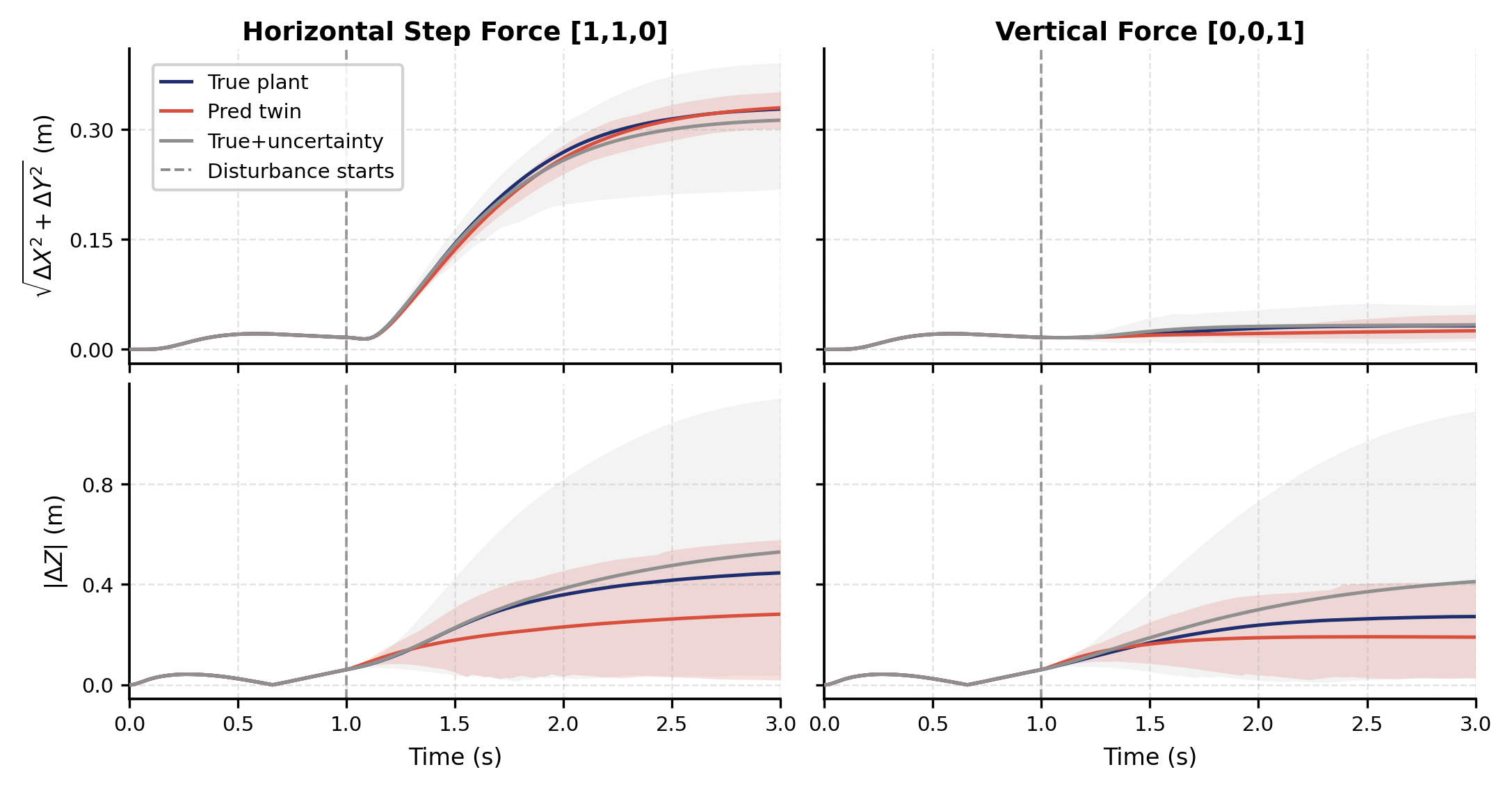}
    \caption{Single-trajectory response to deterministic constant force steps on the nominal hardware-scale system for horizontal force $[1,1,0]\,\mathrm{N}$ and vertical force $[0,0,1]\,\mathrm{N}$. The top row reports lateral error $\sqrt{\Delta X^2+\Delta Y^2}$ and the bottom row reports vertical error $|\Delta Z|$. In each panel, we compare the true closed-loop rollout, decoded-model rollouts sampled from the latent posterior, and a true-system uncertainty baseline (uniform $\pm 20\%$ parameter perturbation) under matched initial conditions and matched disturbance inputs; shaded regions indicate the 5--95\% confidence interval. The decoded-model envelope brackets the true lateral response at the correct scale and is far tighter than the $\pm 20\%$ baseline, while the vertical response is less well captured, localizing the decoder's reliability to the lateral channel.}
    \label{fig:single-traj-step-disturbance}
\end{figure}
We first evaluate deterministic step responses from hover, using the nominal parameters in Table~\ref{tab:quad-params}. We apply either a constant horizontal force $[1,1,0]\,\mathrm{N}$ or a constant vertical force $[0,0,1]\,\mathrm{N}$. For each disturbance, we run one true-system rollout and $100$ decoded-model rollouts from the same latent, all with identical initial conditions and identical disturbance profiles. To separate decoder-induced uncertainty from policy-induced behavior, we also include a baseline that rolls out true-system dynamics with independently sampled uniform $\pm 20\%$ parameter perturbations.

Figure~\ref{fig:single-traj-step-disturbance} shows that the decoded models provide a useful uncertainty envelope for the step response, especially in the lateral direction. For the horizontal force, the true lateral error stays inside the predicted envelope over the rollout. For the vertical force, the lateral error remains small and is also predicted at the correct scale. The vertical-error panels are less accurate: the decoded rollouts have a wider spread and a larger offset from the true trace. Thus, on this individual trajectory, the decoded models capture the lateral closed-loop response better than the vertical response. The rollouts from the true system with $\pm 20\%$ parameter perturbations show a much larger envelope in both directions, indicating that CFM rollouts stay close to the true response because of the decoded models, not because of the policy alone.
\subsubsection{Monte-Carlo Disturbance Analysis}

\begin{figure}[t]
    \centering
    \includegraphics[width=\linewidth]{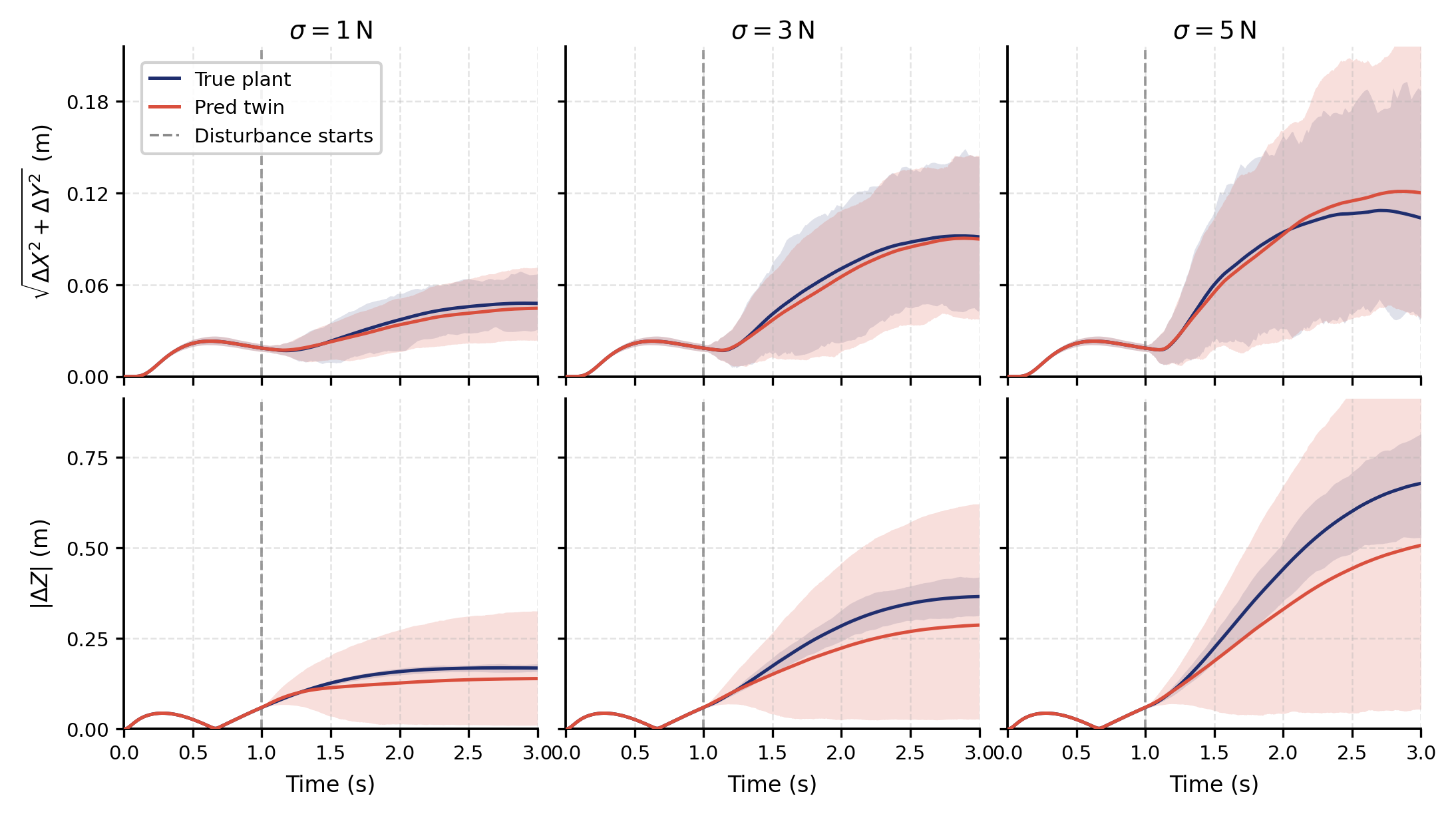}
    \caption{Temporal tracking-error envelopes under i.i.d.\ Gaussian force disturbances with per-axis standard deviation $\sigma=1\,\mathrm{N}$, $3\,\mathrm{N}$, and $5\,\mathrm{N}$. The top row reports lateral error $\sqrt{\Delta X^2+\Delta Y^2}$ and the bottom row reports vertical error $|\Delta Z|$; shaded regions indicate the 5--95\% confidence interval. The decoded envelopes track the growth of the true lateral error as $\sigma$ increases, providing a calibrated robustness envelope in the lateral channel.}
    \label{fig:dist_temporal_lateral_sigma_sweep}
\end{figure}

\begin{figure}[t]
    \centering
    \includegraphics[width=\linewidth]{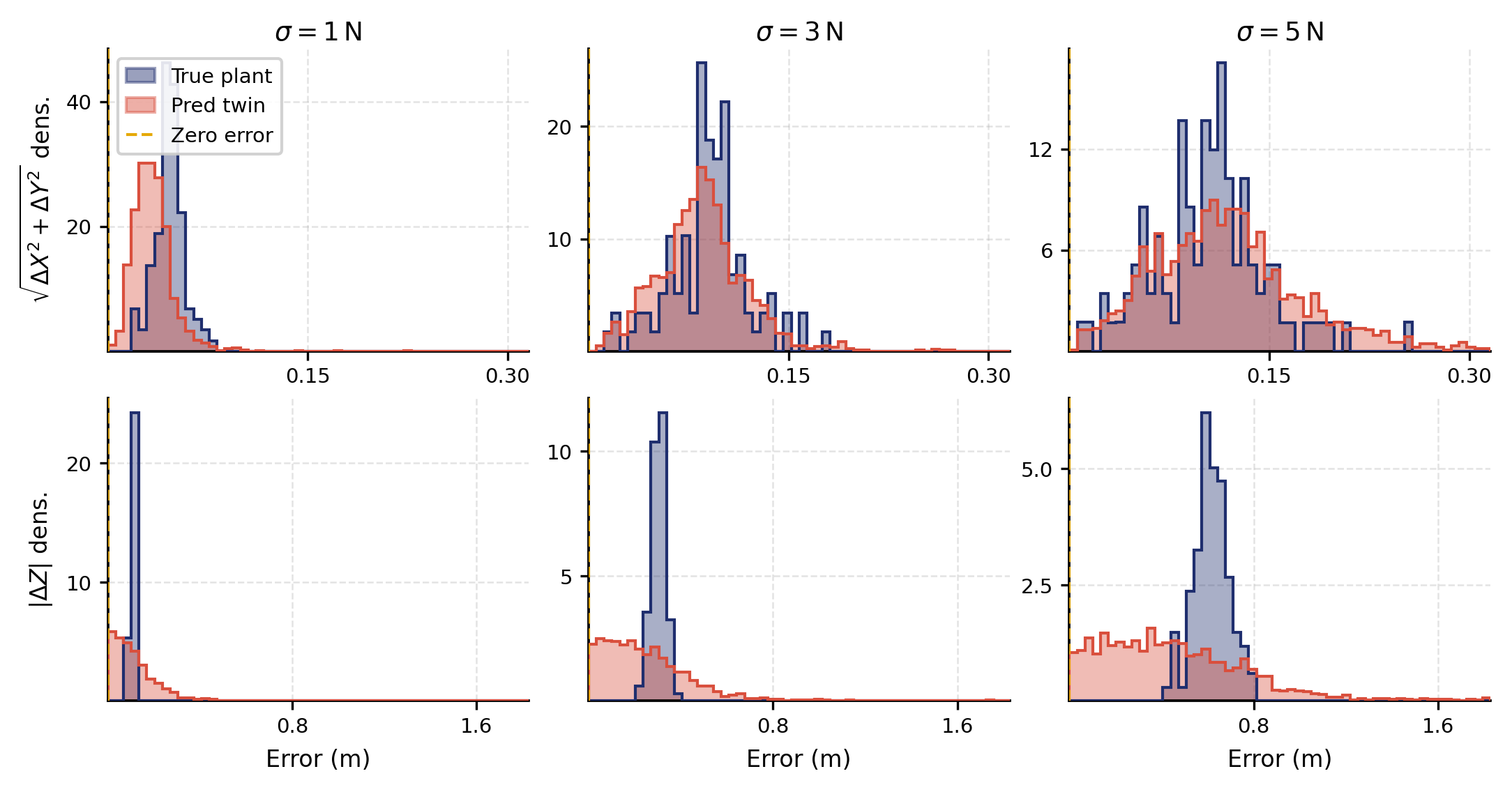}
    \caption{Final tracking-error distributions at the end of the Gaussian disturbance rollouts. Columns correspond to $\sigma=1\,\mathrm{N}$, $3\,\mathrm{N}$, and $5\,\mathrm{N}$; rows show lateral and vertical error. The decoded final-error distributions closely match the true lateral distribution across all $\sigma$, while overestimating the vertical spread, consistent with the temporal envelopes.}
    \label{fig:dist_final_position_sigma_sweep}
\end{figure}

We then test random disturbances, where the exact force realization is unknown but its distribution is specified. The force disturbance is i.i.d.\ zero-mean Gaussian on all three axes, with per-axis standard deviation $\sigma \in \{1,3,5\}\,\mathrm{N}$. For each $\sigma$, we run $N_{\mathrm{MC}}=100$ real-system rollouts in simulation. For every rollout, the decoded-model rollouts use the same initial condition and the same sampled disturbance sequence. We sample $20$ decoded models per rollout, giving $2000$ predicted rollouts per disturbance level.

We report two complementary views. Figure~\ref{fig:dist_temporal_lateral_sigma_sweep} shows the temporal error envelope, which is the more relevant quantity for robustness monitoring because it indicates whether the tracking error may exceed a tolerance at any time during the rollout. Figure~\ref{fig:dist_final_position_sigma_sweep} summarizes the final tracking-error distribution, which is closer to a finite-horizon steady-offset measure. In both figures we separate lateral and vertical errors.

Across the three disturbance levels, the decoded lateral-error envelopes increase as $\sigma$ increases, matching the trend of the true rollouts in Figure~\ref{fig:dist_temporal_lateral_sigma_sweep}. The final lateral-error histograms in Figure~\ref{fig:dist_final_position_sigma_sweep} also remain close to the true distributions. The vertical-error plots show a different behavior: the decoded distribution is broader and shifted relative to the true distribution, especially at larger $\sigma$. This repeats the pattern from the deterministic step test. The decoded models identify that vertical error grows under stronger forcing, but they overestimate or misplace part of that vertical spread.

These experiments show two uses of the decoded ensemble. First, it provides a useful lateral robustness envelope: for both deterministic and Gaussian disturbances, the predicted lateral error has the correct scale and trend, and the deterministic step test rules out that the agreement comes only from Monte Carlo aggregation. Second, the analysis exposes a specific weakness, since the vertical response is less reliable than the lateral one, pointing to where the training distribution or the decoded parameter set should be strengthened.

    \section{Ablation of Distributional Decoding}
\label{sec:exp}
This section tests whether the latent-to-parameter map should be modeled as a distribution rather than a single point estimate. We compare CFM with a deterministic regression baseline that uses the same latent conditioning architecture but predicts one parameter vector. This ablation targets the main modeling choice in our decoder: whether a latent should map to $p(\theta \mid \hat{\bm z}_t)$ or to a single $\hat\theta$.

The neural policy does not explicitly receive physical parameters; it receives only the operational latent, but its motor commands change with that latent. Action replay measures whether decoded parameters can reproduce this latent-to-action map. We fix a recorded hardware trajectory and reuse the same state, command, and latent history for every decoder under comparison. At each time step, the decoder produces a parameter sample $\hat\theta$ that instantiates a parametric controller $\mathcal{K}_{\hat\theta}$ (defined in Appendix~\ref{app:param-ctrl}), and we compare its motor command with the command issued by the latent-conditioned policy on the original rollout. Here $\mathcal{K}_{\hat\theta}$ serves as a probe that exposes $\hat\theta$ in action space, not as a competing controller. Because the procedure is identical across decoders and the only varying input is how $\hat\theta$ is produced from the latent, the resulting motor-command similarity reflects how well each decoder produces
parameters whose induced controller behavior matches the latent-conditioned policy.

\subsection{Action Prediction on Hardware}
\label{sec:action-pred}

\subsubsection{Lemniscate Tracking}
\begin{figure*}[t]
    \centering
    \includegraphics[width=\linewidth]{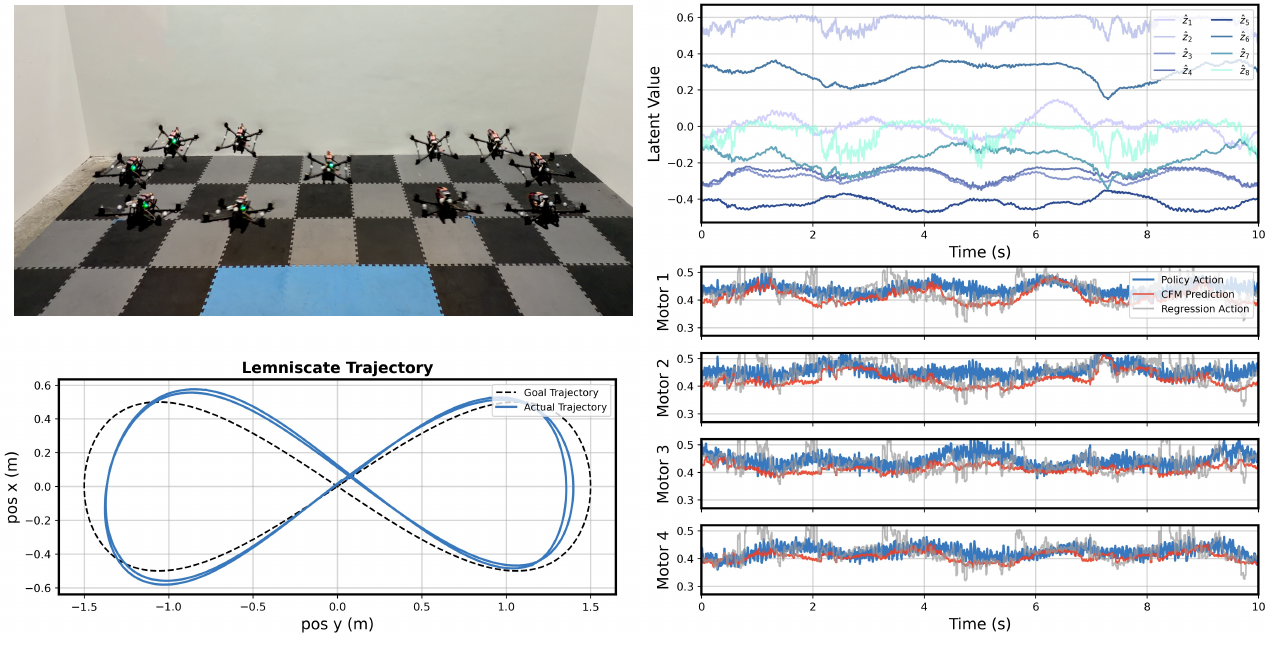}
    \caption{Lemniscate tracking and open-loop action prediction on hardware. The quadrotor completes one lemniscate cycle in $5\,\mathrm{s}$ with position tracking RMSE $0.121\,\mathrm{m}$. The right panels show the latent evolution and normalized motor commands from the neural policy, the CFM-instantiated parametric controller, and the regression baseline. The policy runs at $500\,\mathrm{Hz}$, while CFM and regression decode the latent at $10\,\mathrm{Hz}$. The CFM-decoded controller reproduces the policy's motor commands in both magnitude and temporal pattern, whereas the regression baseline oscillates around them, indicating that the decoded parameters capture the policy's latent-to-action mapping even under aggressive flight.}
    \label{fig:fig8}
\end{figure*}

We evaluate action replay in two hardware settings: aggressive lemniscate tracking and an abrupt off-center payload. For each recorded trajectory, the adaptation module produces the operational latent $\hat{\bm z}_t$ from the measured state history, CFM decodes a parameter sample $\hat\theta$, and the resulting parametric controller $\mathcal{K}_{\hat\theta}$ predicts normalized motor commands. These predictions are compared with the commands issued by the latent-conditioned policy on the same recorded trajectory. Because both controllers are evaluated on the same state, command, and latent history, the comparison isolates action-level consistency between the policy and the decoded dynamics.

We test three lemniscate tracking trajectories, with one cycle completed in $5\,\mathrm{s}$, $10\,\mathrm{s}$, and $15\,\mathrm{s}$. Figure~\ref{fig:fig8} shows the most aggressive $5\,\mathrm{s}$ case. Both the neural policy and the CFM-decoded controller output normalized motor speeds $\Omega/\Omega_{\max}$ at $500\,\mathrm{Hz}$. CFM updates the decoded parameters at $10\,\mathrm{Hz}$ to instantiate $\mathcal{K}_{\hat\theta}$.

We include two baselines. \textit{Random} instantiates $\mathcal{K}_{\hat\theta}$ with a parameter vector obtained by independently perturbing each entry of the measured nominal $\theta$ in Table~\ref{tab:quad-params} by $\mathcal{U}(-50\%,+50\%)$; a fresh sample is drawn at every CFM update step. \textit{Regression} predicts a single point estimate of $\theta$ from the latent using the same conditioning architecture as CFM but without distributional generative modeling, so that the baseline isolates the effect of treating the latent-to-parameter map as a distribution rather than a deterministic function. Table~\ref{tab:action_prediction} reports the root-mean-square error (RMSE) of raw normalized motor commands and the Pearson correlation coefficient (PCC) of action histories.

\begin{table}[t]
    \centering
    \caption{Open-loop action prediction performance on hardware rollouts. RMSE is computed on raw normalized motor commands; PCC is computed after low-pass filtering the action histories. Arrows indicate the preferred direction: $\downarrow$ lower is better (RMSE), $\uparrow$ higher is better (PCC). Best value per row in bold.}
    \label{tab:action_prediction}
    \begin{tabular}{llcc}
    \toprule
    \textbf{Task} & \textbf{Method} & \textbf{RMSE}$\downarrow$ & \textbf{PCC}$\uparrow$ \\
    \midrule
    \multirow{3}{*}{Tracking ($5\,\mathrm{s}$)} 
    & \textit{Random}     & 0.141 & 0.202 \\
    & \textit{Regression} & 0.059 & 0.077 \\
    & CFM        & \textbf{0.032} & \textbf{0.723} \\
    \midrule
    \multirow{3}{*}{Tracking ($10\,\mathrm{s}$)} 
    & \textit{Random}     & 0.151 & 0.080 \\
    & \textit{Regression} & 0.064 & 0.145 \\
    & CFM        & \textbf{0.034} & \textbf{0.710} \\
    \midrule
    \multirow{3}{*}{Tracking ($15\,\mathrm{s}$)} 
    & \textit{Random}     & 0.154 & 0.103 \\
    & \textit{Regression} & 0.073 & 0.172 \\
    & CFM        & \textbf{0.032} & \textbf{0.775} \\
    \midrule
    \multirow{3}{*}{Payload} 
    & \textit{Random}     & 0.156 & 0.205 \\
    & \textit{Regression} & 0.064 & 0.614 \\
    & CFM        & \textbf{0.031} & \textbf{0.812} \\
    \bottomrule
    \end{tabular}
    \end{table}

\begin{figure}
    \centering
    \includegraphics[width=\linewidth]{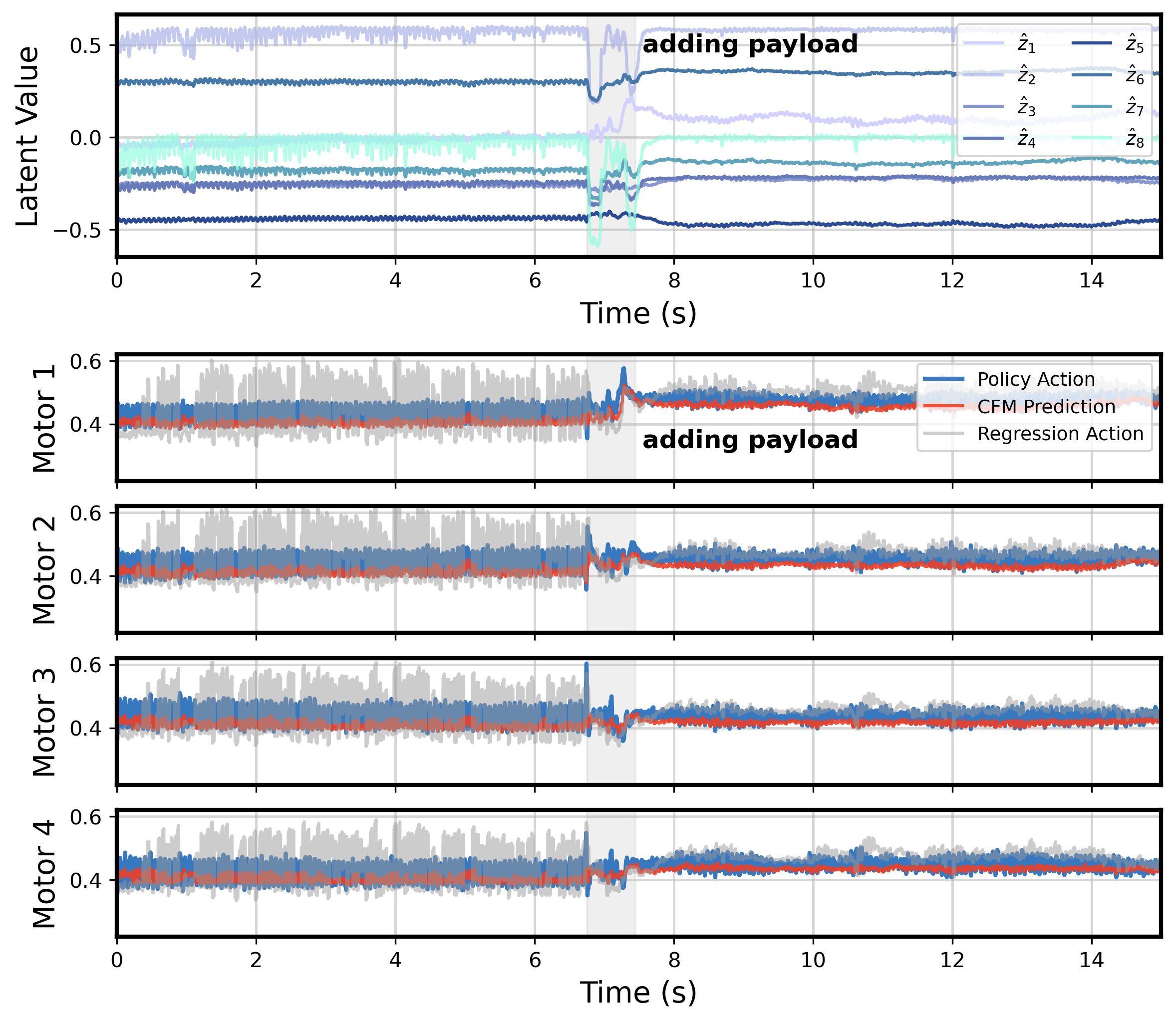}
    \caption{Off-center payload experiment. A $100\,\mathrm{g}$ payload is attached during hover at position $[0.1,\,0.0,\,0.0]\,\mathrm{m}$ in the body frame; the shaded interval marks the attachment event. The upper panel shows latent evolution, and the lower panels compare normalized motor commands from the neural policy, the CFM-instantiated parametric controller, and the regression baseline. When the payload attaches, the latent shifts abruptly and the CFM-decoded commands continue to match the policy through the transient while the regression baseline does not, showing that the decoder follows a sudden physical change and stays consistent with the policy.}
    \label{fig:payload}
\end{figure}

Across all three tracking speeds, CFM gives the lowest motor-command RMSE and the highest PCC. On the $5\,\mathrm{s}$ trajectory, CFM reaches RMSE $0.032$ and PCC $0.723$, compared with $0.059$ and $0.077$ for \textit{Regression}. The regression baseline reduces raw command error relative to \textit{Random}, but its low PCC indicates that the predicted actions do not follow the policy's temporal structure. The gap is largest on the aggressive $5\,\mathrm{s}$ trajectory and shrinks at slower speeds, matching the distributional view: when the latent is strongly excited, the conditional $p(\theta \mid \hat{\bm z}_t)$ is broad because the latent-to-parameter map is not fully identifiable, so the regression target is not a well-defined function. A deterministic regressor is then forced to fit a conditional average that varies erratically as the latent moves and need not correspond to any single self-consistent system. \textit{Regression}'s sensitivity to latent variations is visible in Figures~\ref{fig:fig8} and~\ref{fig:payload}, where the regression traces oscillate around the policy's smoother response. CFM instead draws coherent samples from the conditional and so reproduces both the magnitude and the temporal pattern of the policy commands, keeping RMSE near $0.03$ and PCC above $0.70$ at every speed.

\subsubsection{Off-Center Payload Adaptation}

We further evaluate action prediction under an abrupt physical change. The vehicle hovers while a $100\,\mathrm{g}$ payload, approximately $10\%$ of the vehicle mass, is attached off-center at $54\%$ of the arm length. This perturbation changes the effective mass and inertia and introduces a torque bias. During replay, the same measured state and latent history are used to decode $\hat\theta$ online and evaluate $\mathcal{K}_{\hat\theta}$.

The payload case gives the strongest action-prediction result in Table~\ref{tab:action_prediction}: CFM achieves RMSE $0.031$ and PCC $0.812$. \textit{Regression} attains a moderate PCC of $0.614$, which is noticeably better than on aggressive tracking and consistent with the near-hover setting exciting the latent less aggressively, but its RMSE remains roughly twice that of CFM, and \textit{Random} stays both less accurate and weakly correlated. With the tracking results, this confirms that the decoded distribution reproduces the policy's motor commands under both nominal flight and an abrupt mass change.

    \section{Conclusion and Limitations}

This work presents a framework for interpreting latent representations in adaptive control. Although latent-conditioned policies adapt across varying dynamics, their physical meaning is typically unclear due to non-identifiability. We address this by decoding learned control latents into distributions of dynamically consistent physical models using conditional flow matching, rather than predicting a single parameter estimate. 

The decoded model distribution supports two downstream uses around the fixed policy. In predictive control, it improves high-level gain tuning from naive initialization and adapts better than fixed gains under a motor-lag stress test. In robustness analysis, decoded-model ensembles yield calibrated lateral tracking-error envelopes under Gaussian force disturbances while localizing a residual mismatch to the vertical thrust channel. Overall, learned control latents can be read as distributions over physically meaningful dynamics that support downstream analysis around the frozen policy.  Action-replay
ablations further show that distributional decoding predicts policy motor
commands more accurately than deterministic regression, supporting the need to
model the latent-to-parameter map as a distribution.

Several limitations remain. The recovered models are sensitive to observation and state-estimation noise, which propagates through the latent and can degrade the decoded parameter distribution. The current validation is limited to a quadrotor platform and a structured parametric model; systems with stronger nonlinearities, contacts, or hybrid dynamics may require richer model classes and controller probes. 

Future work will use CFM decoding as an active probing loop for improving the adaptive policy. Starting from a learned latent, we can perturb or stretch the latent representation and run the policy in decoded-model rollouts until the closed-loop response fails or leaves a desired robustness envelope. We can then decode those failure latents back into physical system distributions, revealing what kinds of dynamics correspond to the weak regions of the policy. Those decoded weak systems can be injected into the training distribution, or used to motivate additional identified parameters, so that the next policy is trained directly on the regimes where the current one fails.

    \balance
    \bibliographystyle{IEEEtran}
    \bibliography{main}

@inproceedings{RMA2021kumar,
  title     = {RMA: Rapid Motor Adaptation for Legged Robots},
  author    = {Kumar, Ashish and Fu, Zipeng and Wen, Deepak and Malik, Jitendra},
  booktitle = {Robotics: Science and Systems (RSS)},
  year      = {2021}
}

@article{pearl2019rakelly,
  author     = {Kate Rakelly and
                Aurick Zhou and
                Deirdre Quillen and
                Chelsea Finn and
                Sergey Levine},
  title      = {Efficient Off-Policy Meta-Reinforcement Learning via Probabilistic
                Context Variables},
  journal    = {CoRR},
  volume     = {abs/1903.08254},
  year       = {2019},
  url        = {http://arxiv.org/abs/1903.08254},
  eprinttype = {arXiv},
  eprint     = {1903.08254},
  bibsource  = {dblp computer science bibliography, https://dblp.org}
}

@inproceedings{UP-OSI2017yu,
  title     = {Preparing for the Unknown: Learning a Universal Policy with Online System Identification},
  author    = {Yu, Wenhao and Tan, Jie and Liu, C Karen and Turk, Greg},
  booktitle = {Robotics: Science and Systems (RSS)},
  year      = {2017}
}

@inproceedings{kumar2022adapting,
  title        = {Adapting rapid motor adaptation for bipedal robots},
  author       = {Kumar, Ashish and Li, Zhongyu and Zeng, Jun and Pathak, Deepak and Sreenath, Koushil and Malik, Jitendra},
  booktitle    = {2022 IEEE/RSJ International Conference on Intelligent Robots and Systems (IROS)},
  pages        = {1161--1168},
  year         = {2022},
  organization = {IEEE}
}

@inproceedings{qi2023hand,
  title        = {In-hand object rotation via rapid motor adaptation},
  author       = {Qi, Haozhi and Kumar, Ashish and Calandra, Roberto and Ma, Yi and Malik, Jitendra},
  booktitle    = {Conference on Robot Learning},
  pages        = {1722--1732},
  year         = {2023},
  organization = {PMLR}
}

@article{zhang2025xadap,
  author   = {Zhang, Dingqi and Loquercio, Antonio and Tang, Jerry and Wang, Ting-Hao and Malik, Jitendra and Mueller, Mark W.},
  journal  = {IEEE Transactions on Robotics},
  title    = {A Learning-Based Quadcopter Controller With Extreme Adaptation},
  year     = {2025},
  volume   = {41},
  number   = {},
  pages    = {3948-3964},
  doi      = {10.1109/TRO.2025.3577037}
}

@inproceedings{zhang2023xadapprev,
  author    = {Zhang, Dingqi and Loquercio, Antonio and Wu, Xiangyu and Kumar, Ashish and Malik, Jitendra and Mueller, Mark W.},
  booktitle = {2023 IEEE International Conference on Robotics and Automation (ICRA)},
  title     = {Learning a Single Near-hover Position Controller for Vastly Different Quadcopters},
  year      = {2023},
  volume    = {},
  number    = {},
  pages     = {1263-1269},
  doi       = {10.1109/ICRA48891.2023.10160836}
}

@inproceedings{mellinger2011minimum,
  author    = {Mellinger, Daniel and Kumar, Vijay},
  title     = {Minimum Snap Trajectory Generation and Control for Quadrotors},
  booktitle = {2011 IEEE International Conference on Robotics and Automation},
  pages     = {2520--2525},
  year      = {2011},
  doi       = {10.1109/ICRA.2011.5980409}
}

@article{deboer2005cem,
  author  = {de Boer, Pieter-Tjerk and Kroese, Dirk P. and Mannor, Shie and Rubinstein, Reuven Y.},
  title   = {A Tutorial on the Cross-Entropy Method},
  journal = {Annals of Operations Research},
  volume  = {134},
  number  = {1},
  pages   = {19--67},
  year    = {2005},
  doi     = {10.1007/s10479-005-5724-z}
}

@article{lipman2022flow,
  title   = {Flow matching for generative modeling},
  author  = {Lipman, Yaron and Chen, Ricky TQ and Ben-Hamu, Heli and Nickel, Maximilian and Le, Matt},
  journal = {arXiv preprint arXiv:2210.02747},
  year    = {2022}
}

@inproceedings{alemi2017deep,
  title     = {Deep Variational Information Bottleneck},
  author    = {Alemi, Alexander A and Fischer, Ian and Dillon, Joshua V and Murphy, Kevin},
  booktitle = {International Conference on Learning Representations (ICLR)},
  year      = {2017}
}

@book{Ljung1999,
  title     = {System Identification: Theory for the User},
  author    = {Ljung, Lennart},
  year      = {1999},
  publisher = {Prentice Hall}
}

@article{lee2020learning,
  author   = {Joonho Lee  and Jemin Hwangbo  and Lorenz Wellhausen  and Vladlen Koltun  and Marco Hutter },
  title    = {Learning quadrupedal locomotion over challenging terrain},
  journal  = {Science Robotics},
  volume   = {5},
  number   = {47},
  pages    = {eabc5986},
  year     = {2020},
  doi      = {10.1126/scirobotics.abc5986}
}

@article{ho2020denoising,
  title   = {Denoising diffusion probabilistic models},
  author  = {Ho, Jonathan and Jain, Ajay and Abbeel, Pieter},
  journal = {Advances in neural information processing systems},
  volume  = {33},
  pages   = {6840--6851},
  year    = {2020}
}

@inproceedings{duan2016rl2,
  title     = {RL$^2$: Fast Reinforcement Learning via Slow Reinforcement Learning},
  author    = {Duan, Yan and Schulman, John and Chen, Xi and Wohlhart, Peter and Abbeel, Pieter and Chen, Pieter},
  booktitle = {arXiv preprint arXiv:1611.02779},
  year      = {2016}
}

@inproceedings{hafner2019dream,
  title     = {Dream to Control: Learning Behaviors by Latent Imagination},
  author    = {Hafner, Danijar and Lillicrap, Timothy and Fischer, Ian and Villegas, Ruben and Ha, David and Lee, Honglak and Glover, James},
  booktitle = {International Conference on Learning Representations (ICLR)},
  year      = {2020}
}

@book{van2012subspace,
  title     = {Subspace identification for linear systems: Theory—Implementation—Applications},
  author    = {Van Overschee, Peter and De Moor, BL0888},
  year      = {2012},
  publisher = {Springer Science \& Business Media}
}

@article{kingma2013auto,
  title   = {Auto-encoding variational bayes},
  author  = {Kingma, Diederik P and Welling, Max},
  journal = {arXiv preprint arXiv:1312.6114},
  year    = {2013}
}

@article{hafner2023mastering,
  title     = {Mastering diverse control tasks through world models},
  author    = {Hafner, Danijar and Pasukonis, Jurgis and Ba, Jimmy and Lillicrap, Timothy},
  journal   = {Nature},
  pages     = {1--7},
  year      = {2025},
  publisher = {Nature Publishing Group}
}

@article{beck2010bayesian,
  title     = {Bayesian system identification based on probability logic},
  author    = {Beck, James L},
  journal   = {Structural Control and Health Monitoring},
  volume    = {17},
  number    = {7},
  pages     = {825--847},
  year      = {2010},
  publisher = {Wiley Online Library}
}

@incollection{peterka1981bayesian,
  title     = {Bayesian approach to system identification},
  author    = {Peterka, V{\'a}clav},
  booktitle = {Trends and Progress in System identification},
  pages     = {239--304},
  year      = {1981},
  publisher = {Elsevier}
}

@misc{zhang2025simulationevaluationsuiterobust,
  title         = {A Simulation Evaluation Suite for Robust Adaptive Quadcopter Control},
  author        = {Dingqi Zhang and Ran Tao and Sheng Cheng and Naira Hovakimyan and Mark W. Mueller},
  year          = {2025},
  eprint        = {2510.03471},
  archiveprefix = {arXiv},
  primaryclass  = {cs.RO},
  url           = {https://arxiv.org/abs/2510.03471}
}

@inproceedings{song2021flightmare,
  title        = {Flightmare: A flexible quadrotor simulator},
  author       = {Song, Yunlong and Naji, Selim and Kaufmann, Elia and Loquercio, Antonio and Scaramuzza, Davide},
  booktitle    = {Conference on Robot Learning},
  pages        = {1147--1157},
  year         = {2021},
  organization = {PMLR}
}

@article{tong2023improving,
  title   = {Improving and generalizing flow-based generative models with minibatch optimal transport},
  author  = {Tong, Alexander and Malkin, Nikolay and Huguet, Guillaume and Zhang, Yanlei and Rector-Brooks, Jarrid and Fatras, Kilian and Wolf, Guy and Bengio, Yoshua},
  journal = {Transactions on Machine Learning Research},
  year    = {2023}
}

@inbook{ioannou2010robust,
  author = {Ioannou, Petros and Baldi, Simone},
  year   = {2010},
  month  = {12},
  pages  = {1-22},
  title  = {Robust Adaptive Control},
  isbn   = {9781315218700},
  doi    = {10.1201/b10384-41},
  publisher = {CRC Press}
}

@book{hovakimyan2010L1,
  title     = {$\mathcal{L}_1$ adaptive control theory: Guaranteed robustness with fast adaptation},
  author    = {Hovakimyan, Naira and Cao, Chengyu},
  year      = {2010},
  publisher = {SIAM}
}

@article{bellman1970structural,
  title   = {On structural identifiability},
  author  = {Bellman, Richard and Åström, Karl Johan},
  journal = {Mathematical Biosciences},
  volume  = {7},
  number  = {3-4},
  pages   = {329--339},
  year    = {1970}
}

@inproceedings{fu2023deep,
  title        = {Deep whole-body control: learning a unified policy for manipulation and locomotion},
  author       = {Fu, Zipeng and Cheng, Xuxin and Pathak, Deepak},
  booktitle    = {Conference on Robot Learning},
  pages        = {138--149},
  year         = {2023},
  organization = {PMLR}
}

@book{Mueller2025Dynamics,
  author    = {Mark W. Mueller},
  title     = {Dynamics and Control of Autonomous Flight},
  series    = {Mathematical Engineering},
  publisher = {Springer Cham},
  year      = {2025},
  edition   = {1},
  pages     = {Xiii + 165},
  isbn      = {978-3-031-91870-4},
  doi       = {10.1007/978-3-031-91871-1},
  url       = {https://doi.org/10.1007/978-3-031-91871-1}
}

@article{elfwing2018sigmoid,
  title     = {Sigmoid-weighted linear units for neural network function approximation in reinforcement learning},
  author    = {Elfwing, Stefan and Uchibe, Eiji and Doya, Kenji},
  journal   = {Neural networks},
  volume    = {107},
  pages     = {3--11},
  year      = {2018},
  publisher = {Elsevier}
}

@inproceedings{adamw,
  title     = {Decoupled Weight Decay Regularization},
  author    = {Loshchilov, Ilya and Hutter, Frank},
  booktitle = {International Conference on Learning Representations},
  year      = {2019}
}

@inproceedings{vaswani2017attention,
  title     = {Attention Is All You Need},
  author    = {Vaswani, Ashish and Shazeer, Noam and Parmar, Niki and Uszkoreit, Jakob and Jones, Llion and Gomez, Aidan N and Kaiser, {\L}ukasz and Polosukhin, Illia},
  booktitle = {Advances in Neural Information Processing Systems},
  volume    = {30},
  year      = {2017}
}

@inproceedings{perez2018film,
  title     = {{FiLM}: Visual Reasoning with a General Conditioning Layer},
  author    = {P{\'e}rez, Ethan and Strub, Florian and de Vries, Harm and Dumoulin, Vincent and Courville, Aaron},
  booktitle = {Proceedings of the AAAI Conference on Artificial Intelligence},
  volume    = {32},
  number    = {1},
  year      = {2018}
}

@article{brito2025reflexive,
  title   = {World Models as Reference Trajectories for Rapid Motor Adaptation},
  author  = {Brito, Carlos Stein and McNamee, Daniel},
  journal = {arXiv preprint arXiv:2505.15589},
  year    = {2025}
}

@article{mao2024piwm,
  title   = {Physically Interpretable World Models via Weakly Supervised Representation Learning},
  author  = {Mao, Zhenjiang and Umasudhan, Mrinall Eashaan and Ruchkin, Ivan},
  journal = {arXiv preprint arXiv:2412.12870},
  year    = {2024}
}

@article{wang2025phys2real,
  title     = {{Phys2Real}: Fusing {VLM} Priors with Interactive Online Adaptation for Uncertainty-Aware Sim-to-Real Manipulation},
  author    = {Wang, Maggie and Tian, Stephen and Swann, Aiden and Shorinwa, Ola and Wu, Jiajun and Schwager, Mac},
  journal   = {arXiv preprint arXiv:2510.11689},
  year      = {2025},
  url       = {https://arxiv.org/abs/2510.11689}
}

@inproceedings{ramos2019bayessim,
  title     = {{BayesSim}: Adaptive Domain Randomization via Probabilistic Inference for Robotics Simulators},
  author    = {Ramos, Fabio and Possas, Rafael Carvalhaes and Fox, Dieter},
  booktitle = {Robotics: Science and Systems (RSS)},
  year      = {2019}
}

@inproceedings{heiden2022probabilistic,
  title        = {Probabilistic Inference of Simulation Parameters via Parallel Differentiable Simulation},
  author       = {Heiden, Eric and Denniston, Christopher E. and Millard, David and Ramos, Fabio and Sukhatme, Gaurav S.},
  booktitle    = {IEEE International Conference on Robotics and Automation (ICRA)},
  year         = {2022},
  organization = {IEEE}
}

@article{zhu2024jcdi,
  title   = {Diffusion Model-based Parameter Estimation in Dynamic Power Systems},
  author  = {Zhu, Feiqin and Torbunov, Dmitrii and Ren, Yihui and Jiang, Zhongjing and Zhao, Tianqiao and Yogarathnam, Amirthagunaraj and Yue, Meng},
  journal = {arXiv preprint arXiv:2411.10431},
  year    = {2024}
}

@InProceedings{sandha2021sim2real,
  title = 	 {Sim2Real Transfer for Deep Reinforcement Learning with Stochastic State Transition Delays},
  author =       {Sandha, Sandeep Singh and Garcia, Luis and Balaji, Bharathan and Anwar, Fatima and Srivastava, Mani},
  booktitle = 	 {Proceedings of the 2020 Conference on Robot Learning},
  pages = 	 {1066--1083},
  year = 	 {2021},
  editor = 	 {Kober, Jens and Ramos, Fabio and Tomlin, Claire},
  volume = 	 {155},
  series = 	 {Proceedings of Machine Learning Research},
  month = 	 {16--18 Nov},
  publisher =    {PMLR},
  url = 	 {https://proceedings.mlr.press/v155/sandha21a.html}
}

@inproceedings{tan2018simtoreal,
  title     = {Sim-to-Real: Learning Agile Locomotion for Quadruped Robots},
  author    = {Tan, Jie and Zhang, Tingnan and Coumans, Erwin and Iscen, Atil and Bai, Yunfei and Hafner, Danijar and Bohez, Steven and Vanhoucke, Vincent},
  booktitle = {Robotics: Science and Systems (RSS)},
  year      = {2018},
  doi       = {10.15607/RSS.2018.XIV.010}
}

    \appendices
    \section{Parametric Controller}
\label{app:param-ctrl}
This appendix details the parametric controller $\mathcal{K}_\theta$ used to instantiate the decoded parameters $\hat\theta$ in Section~\ref{sec:action-pred}. Given a physical parameter vector $\theta$, $\mathcal{K}_\theta$ maps a high-level reference to rotor-speed commands via inertia-weighted rate feedback and a rigid-body thrust/torque allocation, following the standard formulation in~\cite{Mueller2025Dynamics,zhang2025simulationevaluationsuiterobust}.

\paragraph{Reference and rate loop.}
The high-level reference is
\begin{equation}
\bm r =
\begin{bmatrix}
a_{\mathrm{cmd}} \\
\bm \omega_{\mathrm{cmd}}
\end{bmatrix}
\in \mathbb{R}^4,
\qquad
a_{\mathrm{cmd}} = T_{\mathrm{cmd}}/m,
\label{eq:app-ref}
\end{equation}
i.e.\ a commanded mass-normalized collective thrust and a commanded body rate. The desired body torque is produced by a proportional rate loop,
\begin{equation}
\bm \tau_{\mathrm{cmd}}
= \bm J\, K_\omega\,(\bm \omega_{\mathrm{cmd}} - \bm \omega),
\label{eq:app-rate}
\end{equation}
where $\bm J$ is the inertia matrix supplied by $\theta$, $\bm \omega$ is the measured body rate, and $K_\omega \in \mathbb{R}^{3\times 3}$ is a diagonal rate-gain matrix shared across platforms.

\paragraph{Thrust/torque allocation.}
Using the allocation matrix $B_{\mathrm{alloc}}(\theta) \in \mathbb{R}^{4\times 4}$, which encodes arm length $l$, thrust coefficient $k_t$, and torque coefficient $k_\tau$, together with the per-rotor efficiency vector $\bm \eta \in \mathbb{R}^4$, the commanded squared rotor speeds are
\begin{equation}
\bm \Omega^2_{\mathrm{cmd}}
=
\frac{1}{k_t}\,
\mathrm{diag}\!\left(\frac{1}{\bm \eta}\right)
B_{\mathrm{alloc}}(\theta)^{-1}
\begin{bmatrix}
m\, a_{\mathrm{cmd}} \\
\bm \tau_{\mathrm{cmd}}
\end{bmatrix}.
\label{eq:app-alloc}
\end{equation}
The per-rotor speed command is $\Omega_{i,\mathrm{cmd}} = \sqrt{(\Omega^2_{\mathrm{cmd}})_i}$, and the normalized command used in the policy action space is $\Omega_{i,\mathrm{norm}} = \Omega_{i,\mathrm{cmd}}/\Omega_{\max}$, matching the policy output defined in Section~\ref{sec:setup}.

Through $(m, \bm J, l, k_t, k_\tau, \bm \eta, \Omega_{\max})$, the map $\mathcal{K}_\theta : \bm r \mapsto \bm \Omega_{\mathrm{norm}}$ depends explicitly on $\theta$, so replacing $\theta$ with a decoded sample $\hat\theta \sim p(\theta \mid \hat{\bm z}_t)$ yields the controller $\mathcal{K}_{\hat\theta}$ evaluated in the experiments of Section~\ref{sec:exp}.

\end{document}